\documentclass[10pt,twocolumn,letterpaper]{article}

\usepackage{wacv}
\usepackage{times}
\usepackage{epsfig}
\usepackage{graphicx}
\usepackage{amsmath}
\usepackage{amssymb}
\usepackage{booktabs}
\usepackage[accsupp]{axessibility}
\usepackage{multirow}
\usepackage{soul}

\def\wacvPaperID{1} 

\wacvapplicationstrack 

\wacvfinalcopy 

\ifwacvfinal
\usepackage[breaklinks=true,bookmarks=false]{hyperref}
\else
\usepackage[pagebackref=true,breaklinks=true,colorlinks,bookmarks=false]{hyperref}
\fi

\begin{document}

    \title{CAD2Render: A Modular Toolkit for GPU-accelerated Photorealistic Synthetic Data Generation for the Manufacturing Industry}

\author{Steven Moonen$^1$, Bram Vanherle$^1$, Joris de Hoog$^2$, Taoufik Bourgana$^2$, \\Abdellatif Bey-Temsamani$^2$, Nick Michiels$^1$ \\
$^1$Hasselt University - tUL - Flanders Make, Expertise Centre for Digital Media\\
$^2$Flanders Make, Gaston Geenslaan 8- B-3001 Leuven, Belgium\\
{\tt\small \{steven.moonen,bram.vanherle, nick.michiels\}@uhasselt.be} \\
{\tt\small \{joris.dehoog,taoufik.bourgana,abdellatif.bey-temsamani\}@flandersmake.be }
}

\maketitle
\thispagestyle{empty}

\begin{abstract}
   The use of computer vision for product and assembly quality control is becoming ubiquitous in the manufacturing industry. Lately, it is apparent that machine learning based solutions are outperforming classical computer vision algorithms in terms of performance and robustness. However, a main drawback is that they require sufficiently large and labeled training datasets, which are often not available or too tedious and too time consuming to acquire. This is especially true for low-volume and high-variance manufacturing. Fortunately, in this industry, CAD models of the manufactured or assembled products are available.
   This paper introduces CAD2Render, a GPU-accelerated synthetic data generator based on the Unity High Definition Render Pipeline (HDRP). CAD2Render is designed to add variations in a modular fashion, making it possible for high customizable data generation, tailored to the needs of the industrial use case at hand. Although CAD2Render is specifically designed for manufacturing use cases, it can be used for other domains as well.
   We validate CAD2Render by demonstrating state of the art performance in two industrial relevant setups. We demonstrate that the data generated by our approach can be used to train object detection and pose estimation models with a high enough accuracy to direct a robot. The code for CAD2Render is available at \url{https://github.com/EDM-Research/CAD2Render}.
\end{abstract}

\section{Introduction}

\begin{figure}
    \centering
    \includegraphics[width=0.32\linewidth]{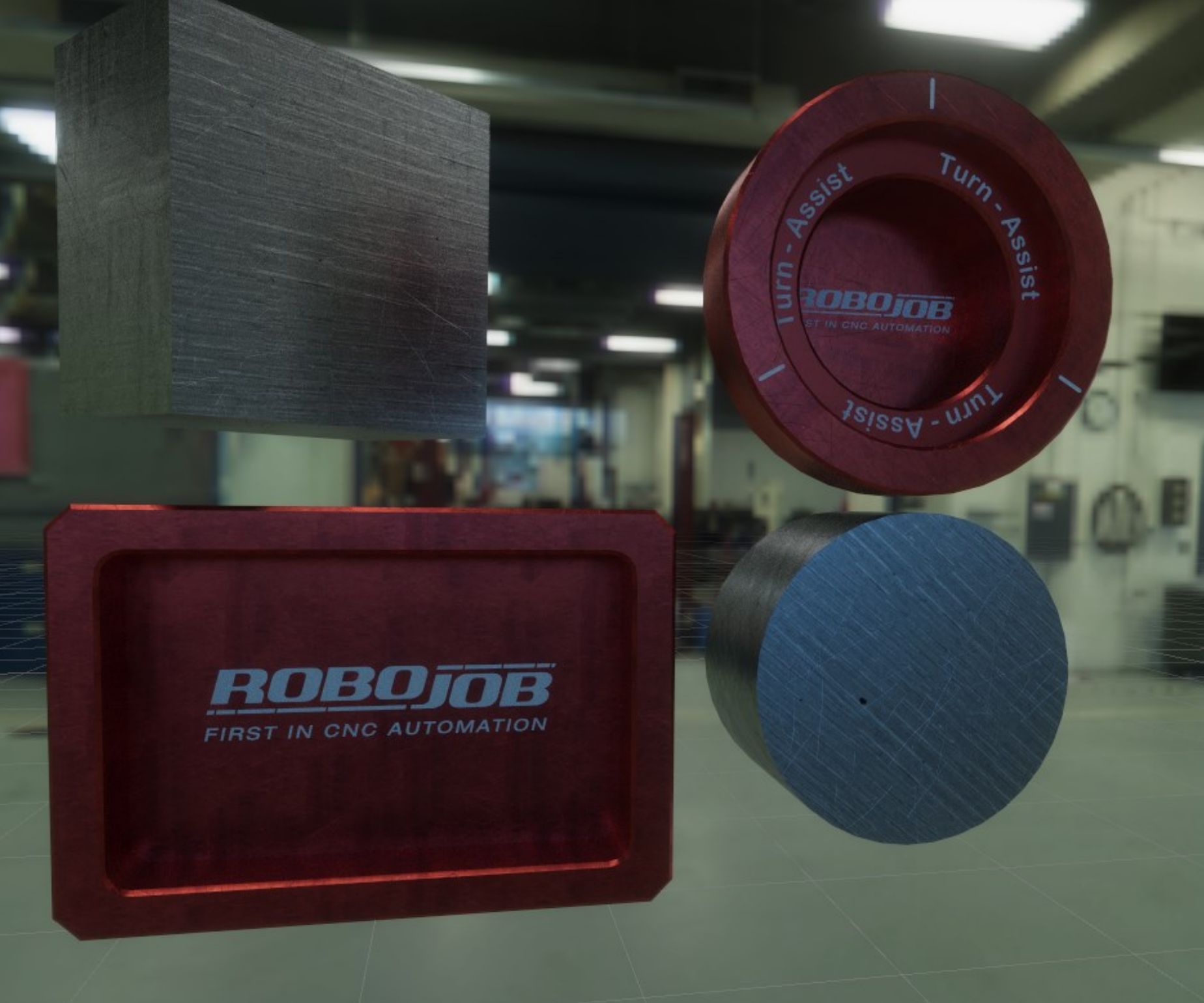}
    \includegraphics[width=0.32\linewidth]{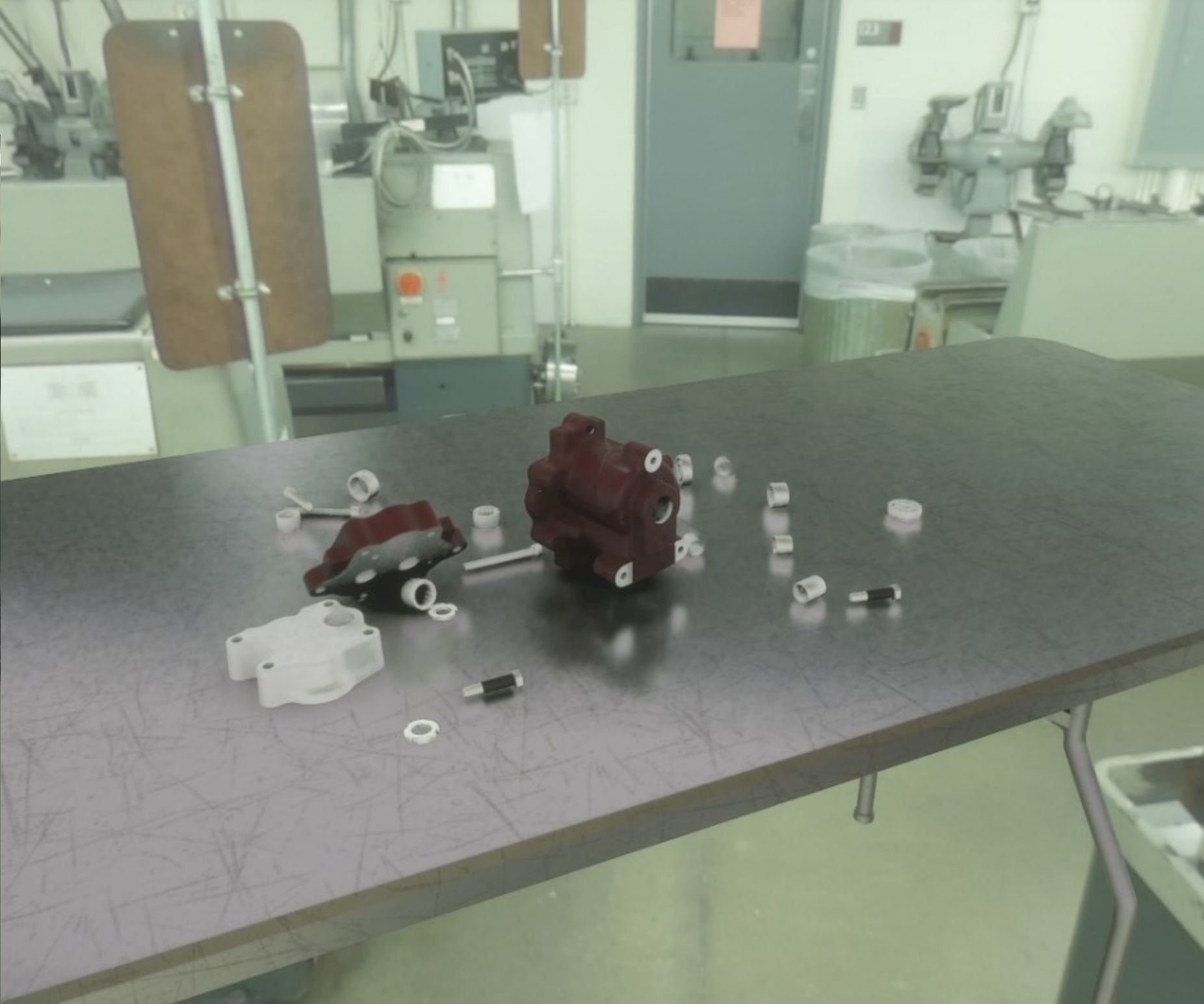}
    \includegraphics[width=0.32\linewidth]{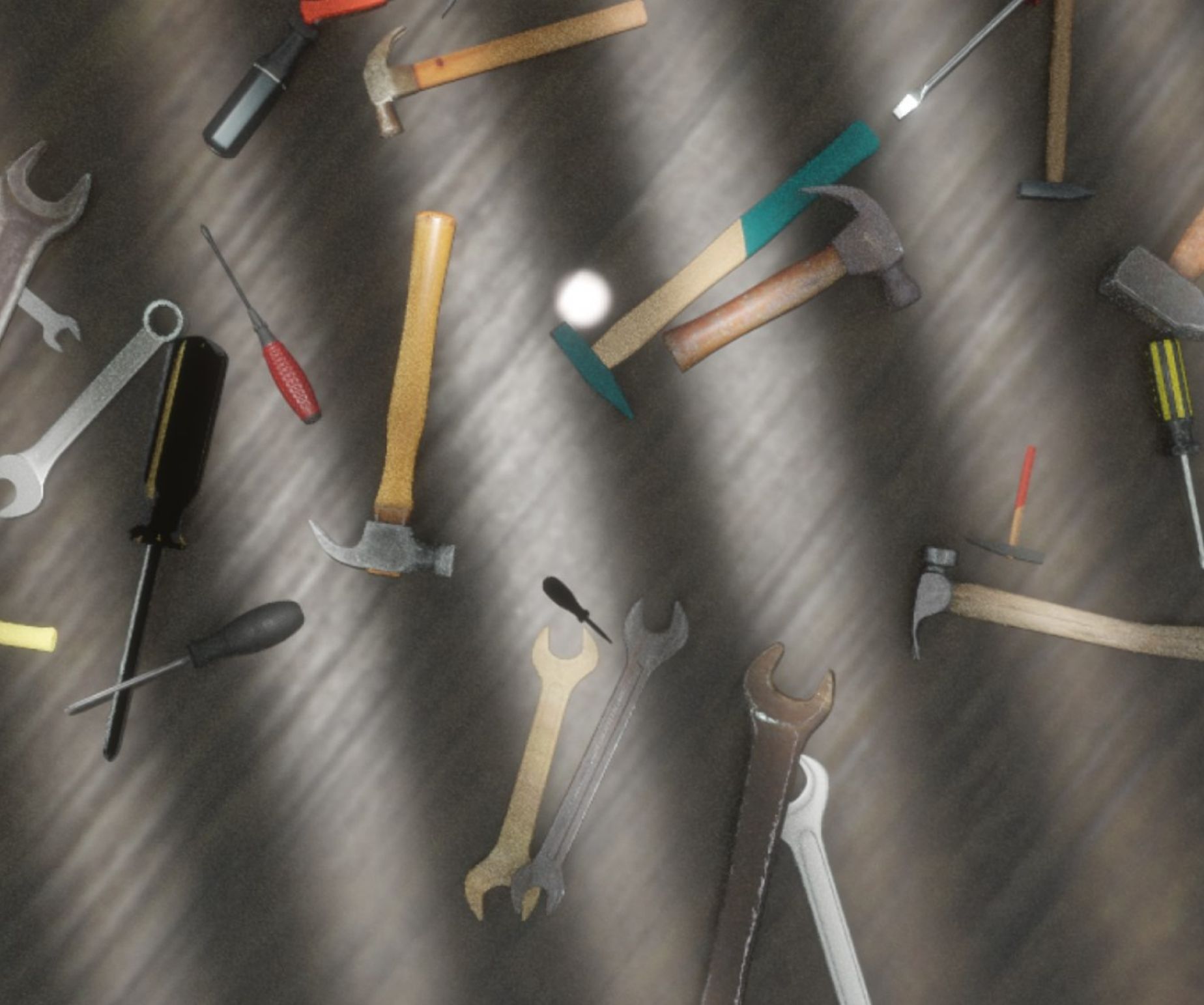}
    \caption{Annotated training set examples of different use cases generated with CAD2Render. left: CNC fabrication, middle: compressor parts, right: tool detection.}
    \label{fig:example_datasets}
\end{figure}
Machine vision has been around in the industrial landscape for a couple of decades and the recent surge in popularity has made the technology a major innovation driver for manufacturers. Most of the approaches are relying on classic vision and are finetuned towards the inspection system. Although they are able to achieve great performance, they require full control of the environment where all the operating conditions stay constant. As such, they are prone to failure when variables like lighting or backgrounds might change. Machine learning algorithms, on the other hand, can be trained to handle such variations, but in turn, have some major challenges that need to be addressed before being fully adopted in this industrial domain.

A first key challenge is the need for large annotated training sets, which are tedious, time consuming and very costly to acquire. In the majority of cases, the data has to be annotated manually, which can lead to bias or errors caused by the human annotator. This limitation is more pronounced in the manufacturing industry, as they have taken some major steps forward in flexible assembly and product manufacturing, allowing them to transform their pipeline towards flexible low-volume and high-variance production. In the extreme case, each produced product can be of a different shape (e.g. prosthesis manufacturing), where there is simply no time to manually capture and annotate datasets. 
\begin{figure*}[t!]
\centering
\includegraphics[width=0.9\linewidth]{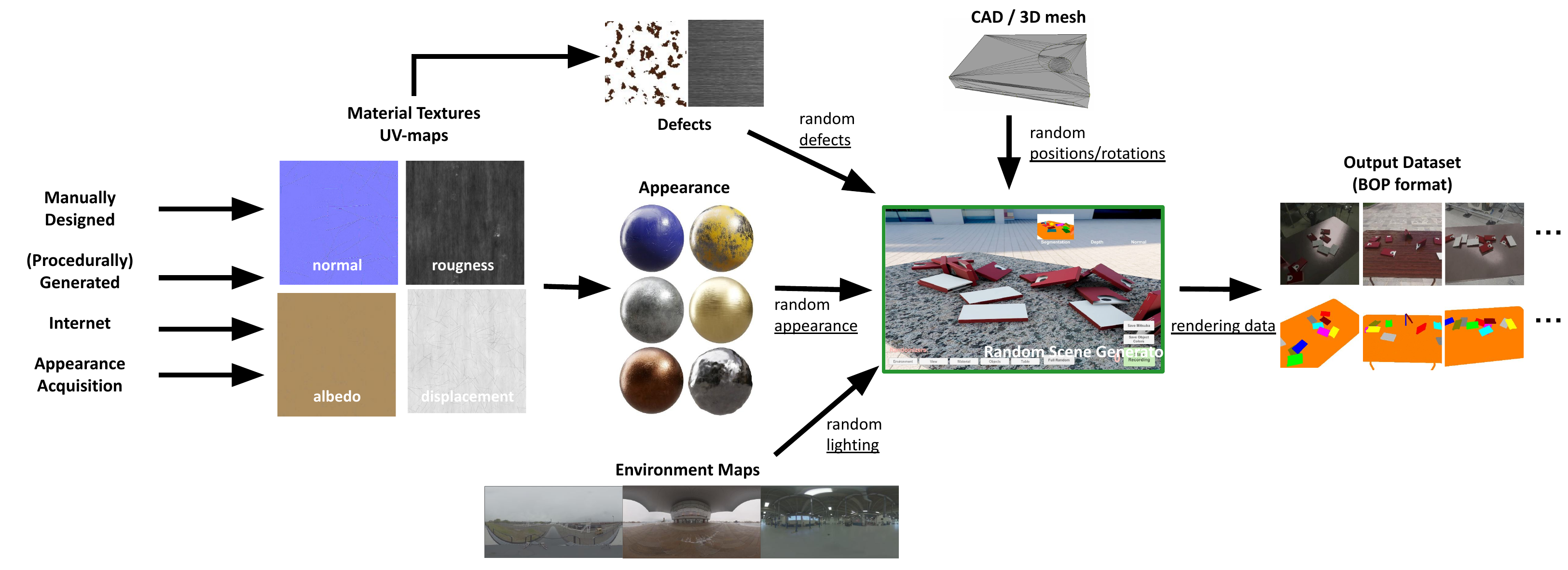}
\caption{Schematic overview of CAD2Render. Environment maps, CAD files and material properties are imported in CAD2Render and used to create a high variety of 3D scene's. Further variations to the material textures can be created by introducing defects like rust or scratches. The scenes are then rendered with a path tracer to create a dataset for training machine learning algorithms.}
\label{fig:CAD2Render}
\end{figure*}
A second challenge is the complex outlook of the materials used in manufacturing. Products are often made of metallic-like materials that cause detailed and complex  reflections. It is very challenging for a vision technique to generalize to all the possible and complex lighting effects.

A common approach to cope with small training datasets is data augmentation: slightly distorting the available data points to create new points that still belong to the same category. While it can provide better performance, it still requires a minimum of data and it does not take into account that the actual products are three-dimensional objects and their visual appearance is governed by complex material properties, lighting conditions, and geometrical detail. This is especially relevant in the manufacturing industry with the metallic-like materials.

This paper exploits the availability of CAD models and the domain knowledge of the manufacturing process, together with the recent advancements in real-time ray tracing, to propose a modular toolkit called CAD2Render. The proposed toolkit is able to automatically generate large amounts of photorealistic training images with extensive visual variations and their accompanying annotations for machine learning purposes. These annotations are generated without human errors or bias. A schematic view of the toolkit is given in Figure~\ref{fig:CAD2Render}. We selected two industrial relevant algorithms for validation, i.e. bin picking by means of object detection and pose estimation, and keypoint detection. Both are important tasks for automating industrial setups, requiring complex vision algorithms. We show that we can achieve a performance capable of solving these tasks in an industrial relevant environment when training on datasets generated by CAD2Render. 

\section{Related work}
Decreasing training set sizes is a popular area of research. Data augmentation is a widely adopted approach to increase the variability in datasets. By slightly distorting the few available images, new examples are created without the need of relabeling the data. The most common variations used are randomly cropping, rotating, scaling, mirroring, color balancing, and adjusting brightness of the entire collection of training pictures to create many slightly modified copies~\cite{7797091, XuJMLCLJ16}. More recently, similar basic image-based variations are generated using deep learning~\cite{abs-1712-04621}. All of these modifications are limited by the information contained in the original 2D pictures, and by the fact that each modification induces a loss of quality. On the other hand, the actual products are three-dimensional and their visual appearance is governed by complex material properties, lighting conditions, and geometrical detail.

Because of the scarcity of training data, recent works propose techniques for training machine learning models purely on synthetic data and have shown that it can achieve similar results compared to SOTA~\cite{Kanade, ShafaeiLS16}. An important conclusion they draw is that realism in the synthetic data is a key factor.
Tobin et al.~\cite{Joshua2017DomRandom} demonstrate the importance of domain randomization when using synthetic data. In their work they show that the domain gap between real and synthetic data can be bridged by introducing enough variations in the synthetic data generation. The machine learning algorithms will see the domain gap as yet another variation of the synthetic data. 

Rendering photorealistic images is a complex task and all the settings for 3D geometry, lighting and materials have to be meticulously modelled in order to achieve convincing photorealism~\cite{Isaac}. In addition, rendering large datasets is a time consuming task, certainly when the generator is based on a ray tracing algorithm to generate the data. This can be a limiting problem because, in low-volume and high-variance manufacturing, new datasets need to be created in a short amount of time. To speed up this task, it can be distributed on a computer cluster or parallelized on a GPU. Kubric created by Greff et al.~\cite{greff2021kubric} is a data generation pipeline that is designed to both work on a single computer to facilitate prototyping or small dataset generation, as well as to run on large computer clusters to speed up the generation. This is only useful when you have access to a computer cluster. 

Jeong et al.\cite{jeong2022perfception} used a NeRF variant to extend a dataset of real images with new viewpoints. This limits the amount of real data required for or a sufficiently large dataset. With NeRF they are able to represent hundreds of photorealistic images in a single format, also reducing the storage size required for the dataset. The data synthesized with this technique can reach better photorealism then rendering a scene from scratch. It also doesn't require an expert optimizing the appearance of a digital scene. However, this technique is not able to add highly customizable variations to the data. 

Other approaches that use 3D rendering focus on a narrow scope of application, such as side view rendering for face detection~\cite{crispell2017dataset}, stereo rendering for depth estimation~\cite{8248284}, vehicle detection~\cite{Alhaija2018,9499331} or large scale factory simulations~\cite{Isaac}. In contrast, most products in industrial manufacturing settings have complex material properties and as a result undergo intricate lighting effects when changing the viewpoint or lighting properties. To include these variations, this paper will focus on algorithms to varying this complex light propagation to cost-effectively synthesize large amounts of training data with accurate and realistic variations of lighting conditions, viewpoints, surface properties, etc. 3D CAD models provided by the manufacturers will be utilized to support this process.

\section{CAD2Render}
CAD2Render is designed as a modular and highly customizable toolkit built upon the HDRP pipeline of Unity3D~\cite{Unity3D} for generating high quality synthetic data for deep learning purposes. It focuses on photorealism by including global illumination effects. A high level overview is provided in Figure~\ref{fig:CAD2Render}. Inspired by the key insight of Tobin et al.~\cite{Joshua2017DomRandom}, that domain randomization is a powerful tool to successfully exploit  synthetic data for training deep learning models, we argue that CAD2Render should support a wide set of complex variations. CAD2Render supports variations such as model types, number of models, instancing, environments lighting, viewpoints, exposure, supporting structures, materials, material appearance, textures, etc. These variations are added in a modular fashion and can be enabled, disabled or extended in function of the use case.

\subsection{Modular Variations}
To facilitate the need for broad and complex variations in the training data, we introduce a wide range of modular randomizers that can introduce different types of variations in the synthetic data. For the clarity of this paper, we have categorized the modules based on pose, lighting, appearance and miscellaneous variations. This section describes how these variations are generated.

\subsubsection{Camera Variations}
The camera pose is randomly defined in spherical coordinates $(\theta,\phi, r)$, in a sphere around a point of interest, orientated  towards this point. The user can define minimum and maximum ranges for these parameters which are then uniformly sampled within this range.
The intrinsic parameters of the camera can also be adjusted to match an existing camera. To further match a physical setup it is also possible to import exact camera poses from a BOP dataset. 

\subsubsection{Object Variations}
The object pose is randomized by automatically spawning new objects in the scene. For each dataset, the user can setup a spawning volume, which defines the 3D region where new objects can be instantiated. Furthermore, the user specifies a ''model path'' that contains the actual models to be spawned, in the form of prefabs. These prefabs can be very simple, just a mesh representation of the CAD model, or can be fully tailored to the use case. CAD2Render will randomly select a set of 3D models from this folder and instantiates them in the scene. The user can specify how many random object are spawned per generated image and if each model is unique or can be instantiated multiple times. In addition, the built-in Nvidia PhysX engine can be enabled or disabled to simulate the objects falling in a natural pose. If enabled, the scene requires a supporting structure, for example a table or pallet. 

\begin{figure}[b!]
\includegraphics[width=\linewidth]{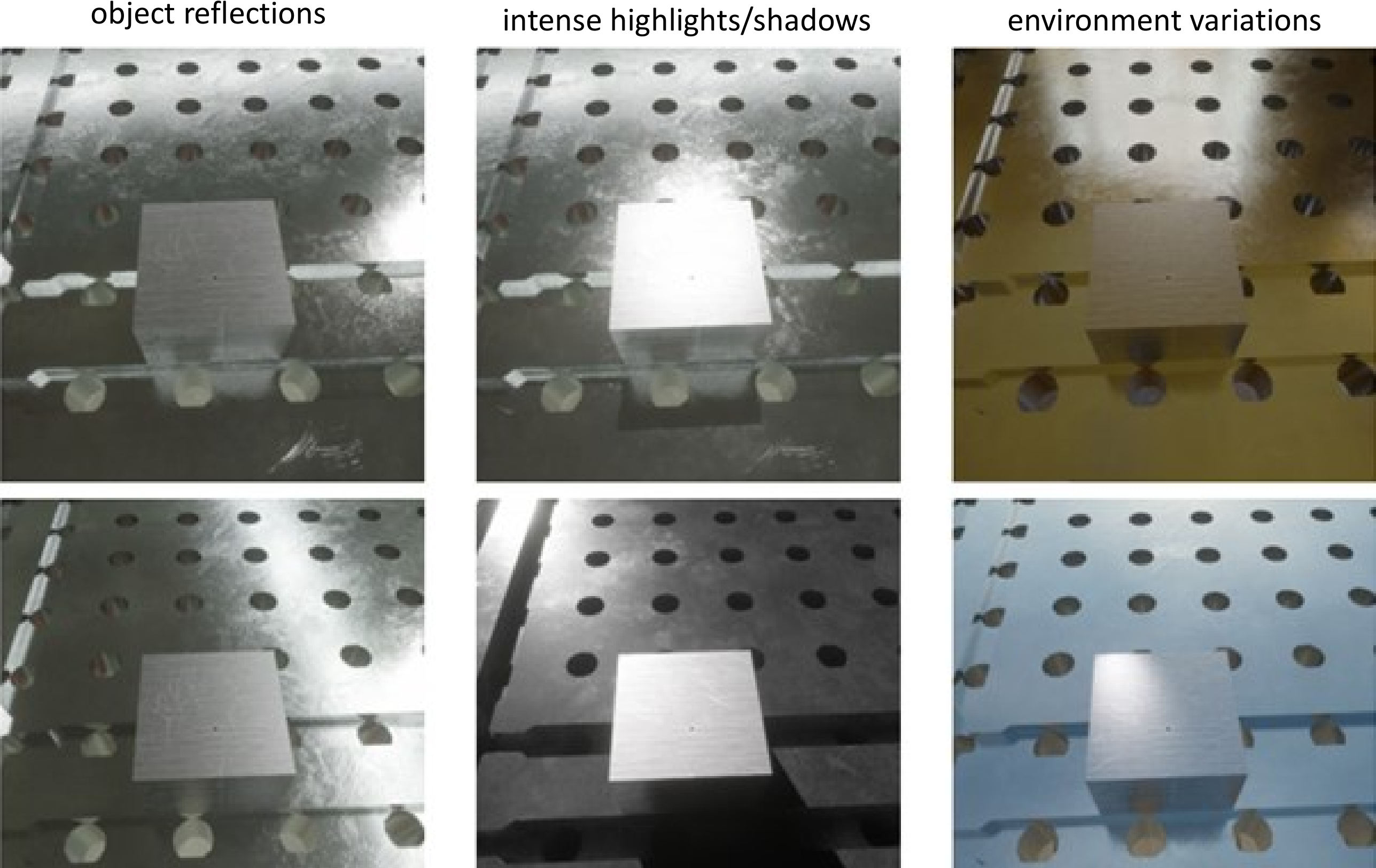}
\caption{Complex lighting variations. Left: inter-reflections between object and and pallet. Middle: intense highlights (top) and hard shadows (bottom). Right: changes in environment lighting. }
\label{fig:lightvariations}
\end{figure}
\subsubsection{Lighting Variations}\label{section:LightingVariations}
Training datasets need sufficient variation in lighting to make deep learning techniques robust for sudden changes in environmental effects~\cite{tremblay_dr}. Two types of variations in light are supported by CAD2Render. 
First, random high dynamic range environment maps are applied to each rendered image. The environment maps are randomly selected from a user specified path. Furthermore, randomized exposure and rotation of the environment map is supported. 
Second, the user can specify additional light source prefabs, acting as templates for additional 3D light sources. During rendering, for each generated image, parameters such as the number or 3D light sources, the intensity, position, rotation and radius can be set and randomized. 

An example of lighting variations is given in Figure~\ref{fig:lightvariations}. The Figure shows examples of environment map variations and variations of intense highlights, shadows and color. We argue that this type of complex photorealistic variations in reflections, shadows and highlights are crucial to incorporate in the training set, because a trained pose or object detector needs to be able to differentiate between what is the actual object and what are the complex light effects that can confuse the model. The more complex light variations the model sees during training, the more it is robust to such changes in a real context. In addition, support for projector variations is available, where the projection of patterns or images can be simulated (example in Figure~\ref{fig:example_datasets} right).

\subsubsection{Appearance Variations}
The appearance of each instantiated object can be varied on-the-fly. The assigned material properties, in the form of normal, roughness, albedo and displacement maps, can originate from different sources.
They can be manually designed, extracted from real sample materials, extracted of the internet or (procedurally) generated (see Figure~\ref{fig:CAD2Render} on the left). In the case of the former three, a database of existing material models can be passed to the CAD2Render toolkit. In the case of the latter, variations of 
 material textures can be automatically generated by the toolkit. At the moment, the toolkit supports three types of texture generators that are industrially relevant: scratches, rust and polishing lines. The settings of these generators can be tailored the the specific needs of the use case at hand. CAD2Render can randomly select and apply materials from a user defined path, similar to the environment maps. Additional variations can be set, with random parameters for HSV offsets, rust, polishing lines, scratches, etc. The database of materials can be provided by the user or can be based on online sources such as the Measured Material Library for Unity HDRP~\cite{MeasuredMaterialLibraryHDRP}.

\begin{figure}[t]
\includegraphics[width=\linewidth]{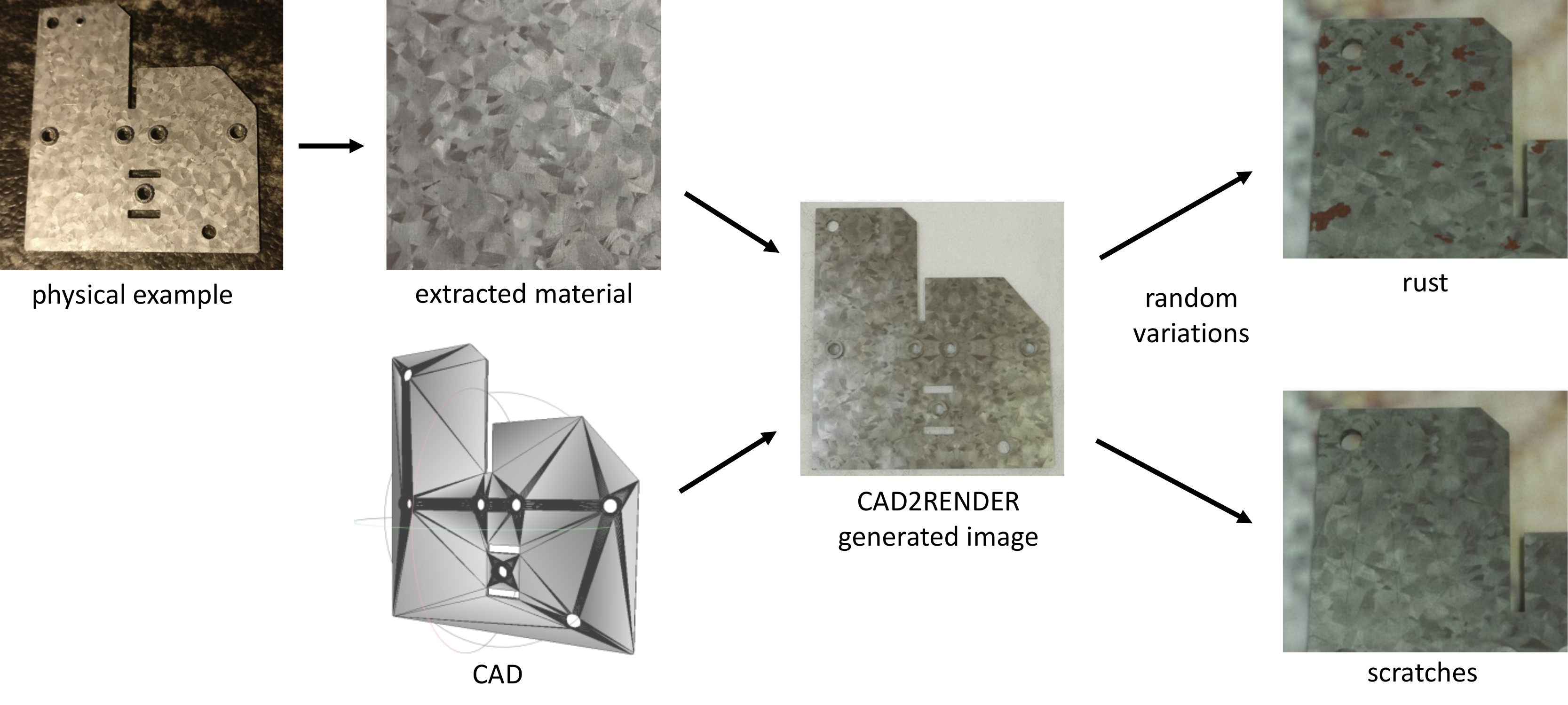}
\caption{Example of rust and scratch variations. An extracted path material from a physical example (top left) is used in combination with a CAD file (bottom left) to generate synthetic images (middle). On top of the extracted material, we can add different material effects such as rust and scratches (right).}
\label{fig:texture_variations}
\end{figure}

The simulation of rust and scratches is inspired by the work of Mihaylov~\cite{Mihaylov2013RustGeneration}. Figure~\ref{fig:texture_variations} shows examples of rust and scratch variations. In this example, we start from a basic material texture, extracted from a real physical example. The extracted material textures are adjusted to include imperfections such as rust or scratches. These variations are generated during the execution with the help of Simplex noise~\cite{Gustavson2005Simplex}. The noise map is used to mark areas where the imperfections need to be generated.
\begin{figure}[b!]
\includegraphics[width=0.90\linewidth]{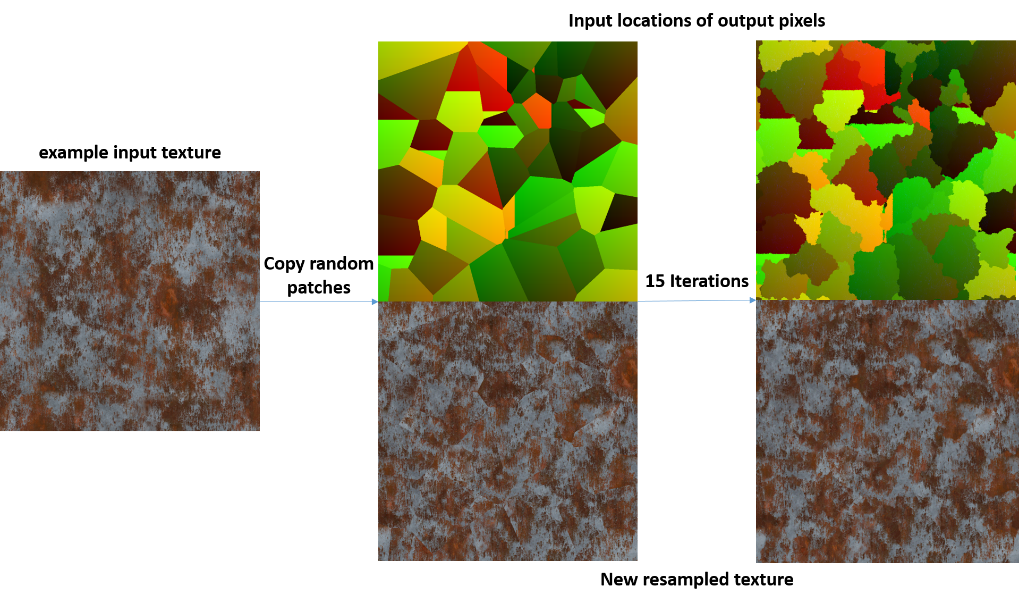}
\caption{Texture resampling. Left: starting material texture. Middle: first iterations of copied patches. Some noticeable hard edges can be perceived. Right: refined result after 15 iterations.}
\label{fig:texture_resampling}
\end{figure}
To allow for more variations in material texture, we have implemented a texture resampling algorithm, based on the work of Opara et al.~\cite{Opara2019Rust}. The goal is to use an input texture and to generate a new texture that looks similar but has some distinct variations. The proposed approach rearranges the pixels of the input texture and tries to limit any obvious seams. The advantage of this technique is that any type of variation can be modeled as long as an example texture is available. Because the technique of Opara et al.~\cite{Opara2019Rust} is designed for offline rendering it is too slow (in the order of minutes) for synthetic data generation. To improve generation time, we have optimized the algorithm for GPU, allowing it to run at interactive frame rates. Figure~\ref{fig:texture_resampling} illustrates the approach of the resampling algorithms. First random patches are copied from an example image (on the left) to the desired output texture. This will create obvious seams where two patches meet (middle part). Then a pixel based algorithm is ran recursively to, step by step, improve the resulting texture (on the right).
Each pixel determines the difference in pixel colors from their current neighborhood compared to the neighborhood they had in the original texture. Then every pixel will calculate the neighborhood difference of multiple pixels suggested by other pixels in the neighborhood and change to the pixel with the lowest difference. This is done multiple times in a row, reducing the radius of the neighborhood with each iteration.

\subsubsection{Miscellaneous}
To conclude, there are some miscellaneous dataset settings that can be set, such as: image resolution, rendering profile (see Section~\ref{sec:acceleration}), post processing profile for white balancing, tonemapping, gamma correction, camera exposure, settings for export, number of physics frames and render frames before export, etc. It is important to note that CAD2Render is built upon Unity and the flexibility of the underlying game engine allows the user to implement or optimize any additional requirements for the use case at hand. This will possibly allow it to be applicable to other domains than the manufacturing industry as well, as long as CAD models are available.

\subsection{Rendering Profiles and GPU Acceleration}\label{sec:acceleration}
The quality and speed of the renders are highly customizable because CAD2Render is based on the High Definition Rendering Pipeline of Unity~\cite{Unity3D}. Unity has a built-in GPU accelerated path tracer that can be used for rendering when photorealism is important. The main drawback of using pathtracing for the generation of the images is the time it takes to render. For some applications a large number of images might be needed. Complementary research has experimented with the amount of CAD2Render images needed to train object detection models and has shown that a large amount of images are beneficial when no domain knowledge is used~\cite{object_detection_synthetic}.

\begin{figure}[b]
\includegraphics[width=\linewidth]{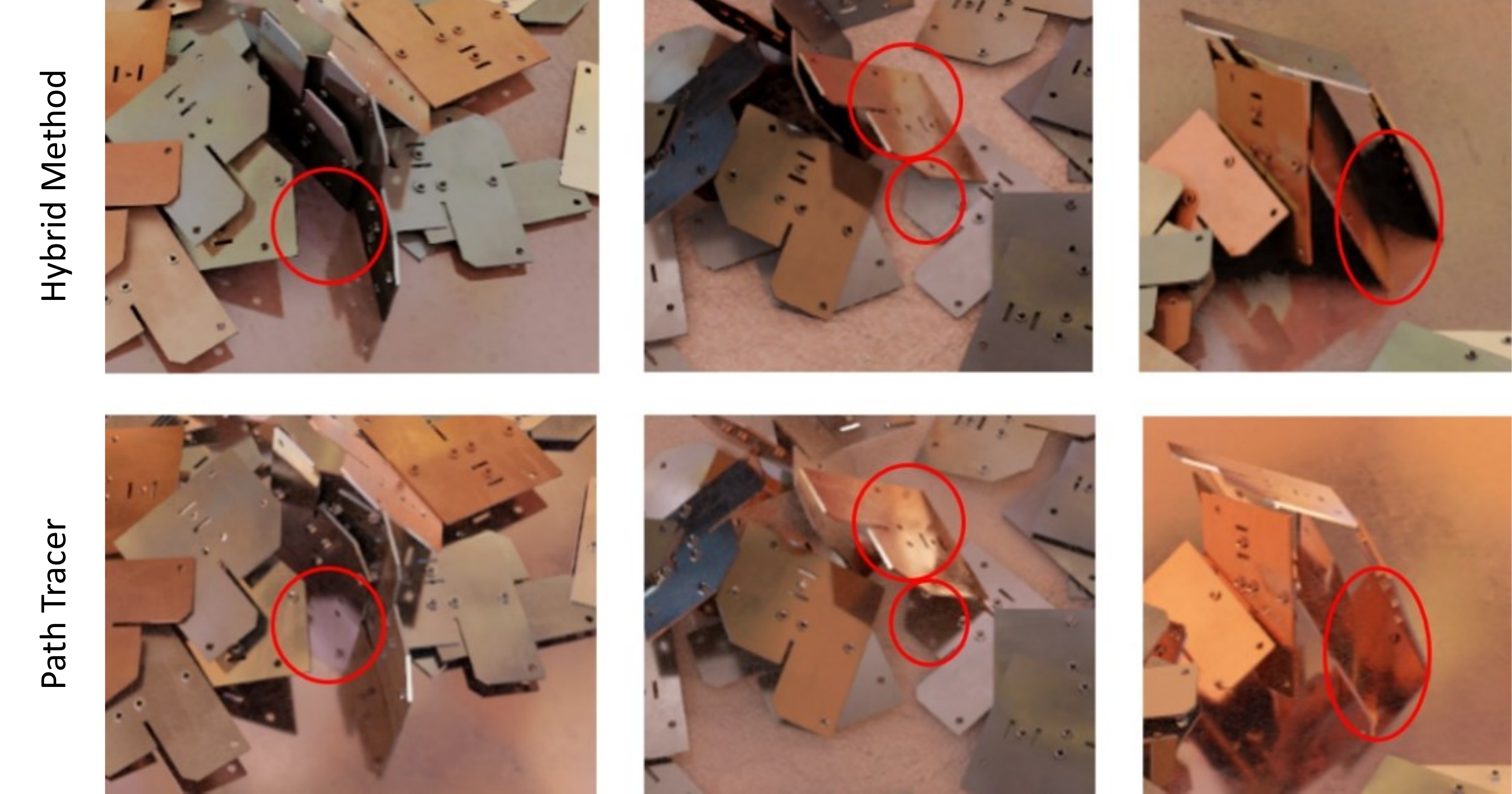}
\caption{Comparison between the Unity path tracer and a hybrid renderer. The Hybrid renderer fails to generate all reflections, shadows and highlights, while the path tracer can introduce noise.}
\label{fig:pathtracing}
\end{figure}

DLSS 2.0 and/or NVIDIA OptiX Denoiser/Intel Open Image Denoiser can be used to reduce the time it takes to generate a converged image. To further increase the generation speed, two other rendering modes can be used: rasterization or hybrid. Rasterization relies on the classic rendering pipeline and hybrid mode is rasterization with limited ray tracing support for shadows, reflections and ambient occlusion.
Changing the rendering approach to rasterization or hybrid can have a considerable positive impact on generation time, but can introduce different types of artifacts. In Figure~\ref{fig:pathtracing} some differences in artifacts are shown between the full pathtracer and the hybrid renderer. The hybrid renderer fails to show all reflections, shadows and highlights that are present in the images generated with the path tracer. The path tracer can introduce noise in areas where the result path tracer converges slowly if no denoiser is used.

Table~\ref{tab:ExecutionTime} gives an overview of the rendering time compared to the resolution between path tracing (at 500 rays per pixel) and rasterization. All measurements were taken with the template scene of the CAD2Render repository on a RTX2070S GPU and an i7-10700KF CPU. Based on empirical observations, enabling the denoiser and rendering $1/10$ of the samples will introduce no noticeable artifacts or noise, allowing for an additional speedup factor of 10. However, further research is required to prove this does not impact the performance of the machine learning models.

\begin{table}[b]
\footnotesize
\centering
\begin{tabular}{|rrrrrr|}
\hline
\multicolumn{6}{|c|}{Rendering 10.000 images}                                                                                                              \\ \hline
\multicolumn{1}{|c|}{Resolution} &
  \multicolumn{2}{c|}{\begin{tabular}[c]{@{}c@{}}Path tracing\\ 500 samples\\ time (hours)\end{tabular}} &
  \multicolumn{2}{c|}{\begin{tabular}[c]{@{}c@{}}Rasterization\\  time (hours) \\\end{tabular}} &
  \multicolumn{1}{c|}{\begin{tabular}[c]{@{}c@{}}\\   Memory   \\ (GB)\end{tabular}} \\ 
  
 \multicolumn{1}{|c|}{} & \multicolumn{1}{|c}{DLSS}  & \multicolumn{1}{c|}{\st{DLSS}} & \multicolumn{1}{|c}{DLSS}  & \multicolumn{1}{c|}{\st{DLSS}} &  \\ \hline
\multicolumn{1}{|r|}{1280 x 720}                                                           & \multicolumn{1}{r|}{7.9} & \multicolumn{1}{r|}{7.8} & \multicolumn{1}{r|}{2.3} & \multicolumn{1}{r|}{2.3} & \multicolumn{1}{r|}{5.4}  \\ \hline
\multicolumn{1}{|r|}{1920 x 1080}                                                          & \multicolumn{1}{r|}{8.3} & \multicolumn{1}{r|}{13.2} & \multicolumn{1}{r|}{2.4} & \multicolumn{1}{r|}{2.4} & 11.3 \\ \hline
\multicolumn{1}{|r|}{3840 x 2160}                                                          & \multicolumn{1}{r|}{22.3} & \multicolumn{1}{r|}{30.0}   & \multicolumn{1}{r|}{3.4}& \multicolumn{1}{r|}{3.6}  & 43.8 \\ \hline
\end{tabular}
\caption{Generation time and storage size for a dataset of 10.000 images. Comparison between various resolutions and render mode (path tracing compared to rasterization).  Measured on a RTX2070 Super GPU and an i7-10700KF CPU at 3.80GHz.}\label{tab:ExecutionTime}
\end{table}

\subsection{Exporting to BOP format}
To allow for easy application of the generated datasets, CAD2Render exports the annotated dataset to the standardized BOP format~\cite{hodan2018bop}. This file format contains RGB images, object and camera poses, camera parameters, instance segmentation, depth maps and 3D models. The BOP format and its accompanying toolkit are originally designed for easy-of-use benchmarking for pose estimation, but due to the rich annotations it can be used for other tasks, such as object detection, segmentation and depth estimation.

\subsection{Importing BOP for digital twin creation}
CAD2Render supports the import of existing BOP datasets as well. This is especially useful for creating a digital twin dataset of a real dataset. This feature makes it easier to research the domain gap between synthetic and real images. The Dataset of Industrial Metal Objects~\cite{dimo} is one such digital twin dataset, generated by CAD2Render.

\section{Validation}
Validation of CAD2Render is done on two industrially relevant use cases: bin picking and 2D keypoint detection. The former requires solutions for both object detection and pose estimation. The latter is a useful approach to detect important landmarks. The validation results are trained solely on synthetic data and tested on real data. Bin picking is applied to metallic objects on a table. Keypoint detection is done for two use cases: tools in use in natural environments and assembly validation on a top down view of a work piece. This wide range of applications highlights the customizability of the toolkit.

\begin{figure}[t]
\centering
\includegraphics[width=0.65\linewidth]{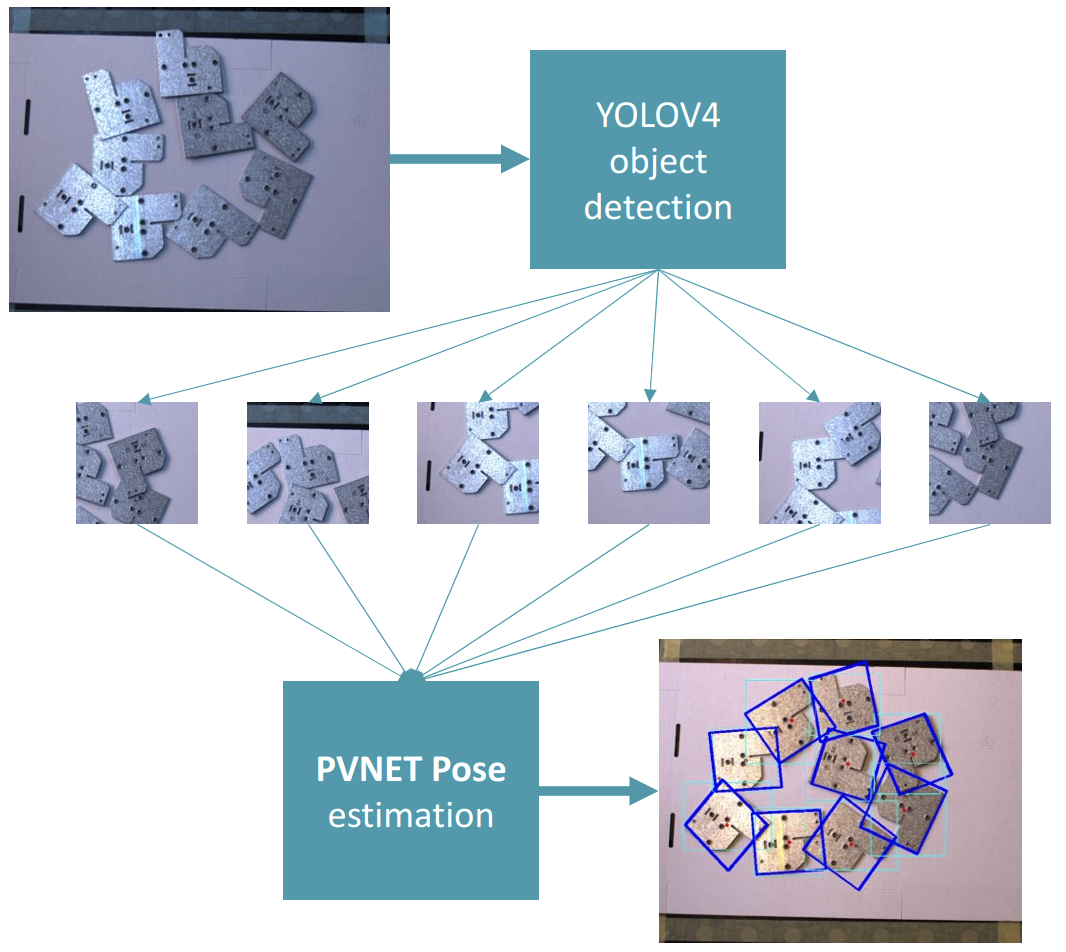}
\caption{Overview of the flow of the validation algorithm}
\label{fig:schematic_overview}
\end{figure}
\subsection{Bin Picking of Metal Objects}
The proposed method has been validated in a setup representing an industrial pick and place application, using a collaborative robot (cobot), equipped with a suction cup and a static camera. The identification of the objects and subsequent position and pose estimation was done with two state of the art networks: YoloV4~\cite{Bochkovskiy2020yolo} for object detection and PVNET~\cite{Peng2019PVNET} for pose estimation. These models are applied sequentially on the input image, where the object detector will calculate the crop in which the object is found. This crop is then used as input for the pose estimator, calculating a full coordinate set for the object in the image. In the case where multiple items are found by the object detector, each crop is processed individually. Figure \ref{fig:schematic_overview} shows a high-level overview of how both networks are used. \\\\
\textbf{Dataset Description} \\
A large dataset of 20.000 images was generated. The images contain from 1 to 10 identical items. The item used for this study is a small stamped metal piece, approximately 7 by 7 centimeter and not symmetrical. The rendered images feature random camera angles, within a defined window of heights and angles. Also, the randomized ambient lighting was applied, as explained in Section~\ref{section:LightingVariations}.\\\\
\textbf{Validation Setup}\\
Quantifying the accuracy of the combined pipeline has to be performed with the appropriate hardware considerations in mind. More precisely, an accurate calibration of the camera pinhole model \cite{pinholecam} is vitally important. For this, a ChArUco board was used, which is a combination of a chequerboard and ArUco markers.
This board is also used to perform the extrinsic calibration, where the relationship between camera pixels and the world geometry is established by calculating the transformation from the camera coordinate system to the world coordinate system. A reference point (or origin point) is established from which geometrical distances can be calculated. 

\begin{figure}[t]
\centering
    \includegraphics[width=0.75\linewidth]{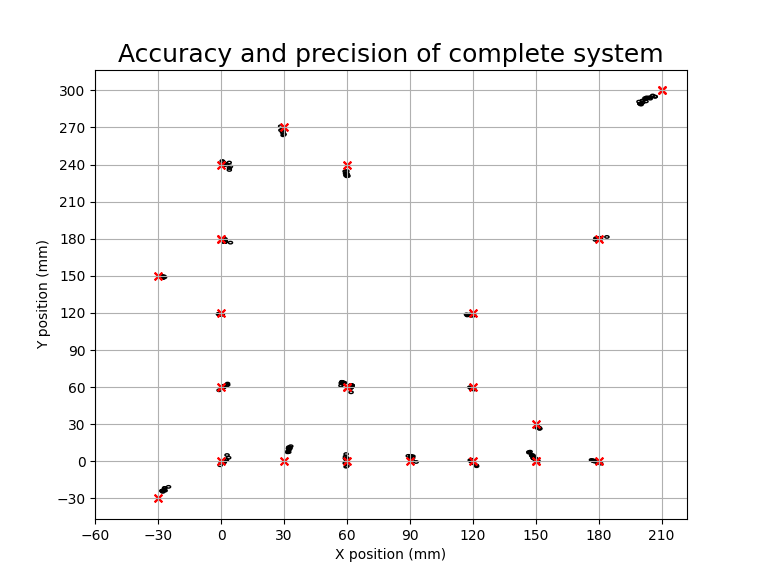}
    \caption{Result of the validation methodology. The red X represents the ground-truth, while the black dots are the results of subsequent estimations 
    by the Pose Estimator. }
    \label{fig:validation_results}
\end{figure}

The validation method then follows the following steps: (1) carefully place the item on the grid; (2) use the pose estimator to estimate the location of a corner of the item (loop for 10 times); (3) calculate geometrical position of this point with respect to the origin point; (4) calculate difference with known location. The camera used in this setup was an IDS UI-3280CP Rev. 2, which has a 2456x2054 pixel sensor and a global shutter. \\\\
\textbf{Validation Results}\\
The object was placed at various locations of the ChArUco board, generally at 30mm intervals. As no rotations of the object were performed, the validation is only valid for the position estimations and not for the rotation estimation. Figure \ref{fig:validation_results} shows the result of the validation methodology explained in this section. It can be seen that there is a specific area in the image outside which the subsequent estimations of the keypoint show relatively large deviations. 

\begin{table}[t]
\footnotesize
\centering
\begin{tabular}{|rrrr|}
\hline
\multicolumn{1}{|c|}{pos} & \multicolumn{1}{c|}{Std Dev {[}mm{]}} & \multicolumn{1}{c|}{Max dev {[}mm{]}} & Min dev {[}mm{]} \\ \hline
\multicolumn{1}{|c|}{x}   & \multicolumn{1}{r|}{0.91}             & \multicolumn{1}{r|}{7.50}             & 0.040             \\ \hline
\multicolumn{1}{|c|}{y}   & \multicolumn{1}{r|}{1.37}             & \multicolumn{1}{r|}{9.92}             & 0.002            \\ \hline
\multicolumn{1}{|c|}{z}   & \multicolumn{1}{r|}{11.16}            & \multicolumn{1}{r|}{57.90}             & 0.220             \\ \hline
\end{tabular}
\caption{Validation Results of the position estimation for Metal Plates on the ChArUco board.}
\label{tab:val_results}
\end{table}

Table \ref{tab:val_results} shows the statistical analysis of the combined results. It can be seen that the standard deviation in the Y-direction is slightly larger compared to the X-direction, which can be explained by the fact that the used camera sensor is not square and therefore there are more pixels in the Y-direction. This means that these pixels fall in the more heavily curved part of the lens, increasing the deviation. 
More generally, it is shown that, provided good camera calibration, the models can be accurate to close to 1 mm in both X and Y directions, which will be more than good enough in most pick-and-place applications. The estimation in the Z-direction is large, however this is a classic issue with height estimations from 2D images that can be improved in the future with a multi-camera setup. \\\\
\textbf{Robustness against harsh light conditions} \\
To demonstrate the robustness of the developed algorithms against harsh lighting conditions, a simple physical setup was conceived on which an object at a known location and orientation could be subjected to either high or low lighting conditions. A dataset was rendered following the method explained in \ref{section:LightingVariations}, this time using a slightly larger metallic object. A simple method of counting the saturated pixels on the surface of the object under test gave an approximate value of light intensity. From Table \ref{tab:lighting_val}, we can see that under all but the most harsh conditions, the XY-deviation, calculated as the Euclidean distance between the ground-truth and the estimated location, is low. It can be observed that when the light intensity reaches more than 70\%, the deviation reaches around 1cm, which can be considered to be extreme.

\begin{table}[h]
 \centering
 \footnotesize
 \begin{tabular}{|r|r|r|r|}
 \hline
 \textbf{Intensity (\%)} 
 & \textbf{XY Deviation {[}cm{]}} \\ \hline
 0.000                    
 & 0.276                          \\ \hline
 35.440                   
 & 0.340                          \\ \hline
 44.090                  
 & 0.742                          \\ \hline
 69.990                   
 & 0.565                         \\ \hline
 73.170                   
 & 1.460                          \\ \hline
 83.660                   
 & 2.444
 \\ \hline
 \end{tabular}
 \caption{Deviation of the estimated XY-location of an object under a wide range of light intensities.}
 \label{tab:lighting_val}
 \end{table}

\subsection{2D Keypoint Detection}
As an additional validation case, CAD2Render was used to train models for the problem of semantic 2D keypoint detection. Specifically, two problems are tackled: localizing keypoints in images of different hand tools and using keypoint detection to validate the assembly of aluminium beams. To find the landmarks, a UNET~\cite{unet} type architecture with intermediate supervision was trained for to generate probability maps for each semantic keypoint location. The problem of tool keypoint detection was investigated by Vanherle et al.~\cite{tool_keypoints_journal} using the CAD2Render tool, for more details consult their paper.

\subsubsection{Tool Keypoint Detection Use Case}
\begin{figure}[h]
    \centering
    \includegraphics[width=0.24\linewidth]{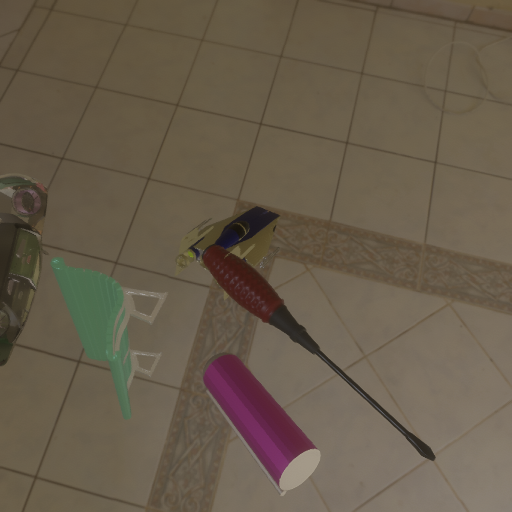}
    \includegraphics[width=0.24\linewidth]{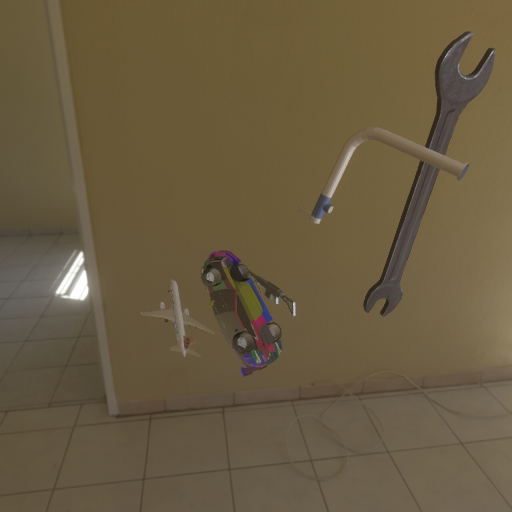}
    \includegraphics[width=0.24\linewidth]{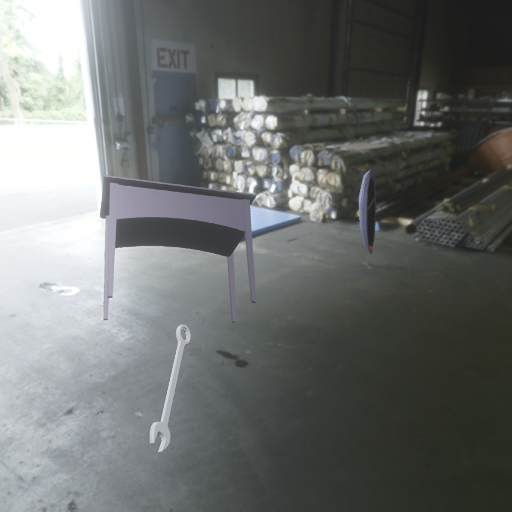}
    \includegraphics[width=0.24\linewidth]{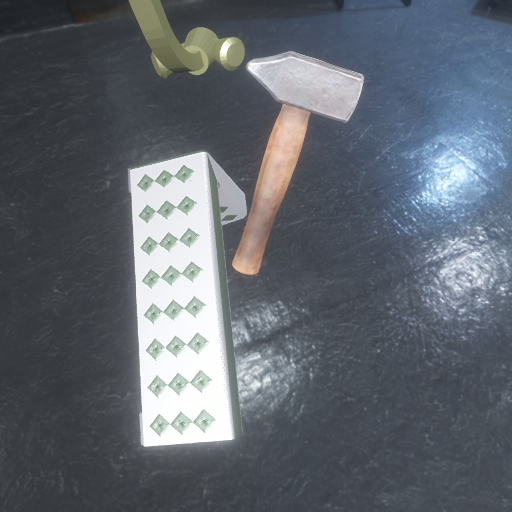}
    \caption{A few examples from the tool keypoint detection dataset created by CAD2Render.}
    \label{fig:tool_images}
\end{figure}

For this use-case we attempt to find the location of certain sementic keypoints of hand tools. The tools considered are a screwdriver, hammer, wrench and combination wrench. For each of these tools we find a number of keypoints by training a model for each tool.\\\\
\textbf{Dataset Description} \\
To train such models, a large amount of data is needed. For each tool we collected a few textured 3D models from the internet. The 3D tools from the internet did not closely resemble the target tools, but did belong to the same class of tool. The CAD2Render toolkit was used to generate 20.000 images for each tool. The tools were randomly spawned in a space with a random environment map as background. Additionally, a few random objects from ShapeNet~\cite{shapenet} were also spawned in the space to simulate occlusions. For this validation case, the faster hybrid renderer was used. Figure~\ref{fig:tool_images} shows some examples of synthetically generated images from the tool keypoint dataset.
\\
\textbf{Validation Setup} \\
To verify whether the datasets generated by CAD2Render are suitable to train a keypoint detection model on, we test the performance of the trained model on real images of tools. For each of the four tools we captured 50 real photographs, and manually annotated these images. To properly test the robustness of the trained models objects were photographed in wide variety of poses, lighting conditions, camera angles and backgrounds. Additionally, occlusions and truncations are introduced. The model trained on the synthetic data was then used to detect keypoint locations in the real images.\\\\

\textbf{Validation Results} \\
To measure the model's performance we use the Percentage of Correct Keypoints~(PCK)~\cite{pck_paper} metric with an $\alpha$ of 1.0. The results are shown in Table~\ref{tab:tool_results}. The models were trained on synthetic images, created randomly without taking domain knowledge into account. Yet, these models are able to detect keypoints in the unseen real images with good accuracy. This shows that the CAD2Render toolkit is able to create synthetic images that are suitable for this problem space and that images produced by the faster hybrid rendering mode can produce good models as well. Additionally, research has shown that models trained on images generated by CAD2Render perform better on this task than models trained on images generated by simple 2D image augmentations~\cite{tool_keypoints_robovis21}. This shows the benefit of using 3D information to generate training data.
\begin{table}[h]
\footnotesize
\centering
\begin{tabular}{|c|l|}
\hline
\textbf{Tool}     & $\textrm{PCK}_{0.1}$ \\ \hline
Screwdriver        &  86.1 \\
Wrench             &  88.9 \\
Combination Wrench &  86.1 \\
Hammer             &  84.4 \\ \hline
\end{tabular}
\caption{Accuracy of the keypoint detection models for each tool trained on the synthetic data. Performance is measured in $\textrm{PCK}_{0.1}$ over the validation set of real images~\cite{tool_keypoints_journal}.}
\label{tab:tool_results}
\end{table}

\subsubsection{Assembly Validation Use Case}
\begin{figure}[t]
    \centering
    \footnotesize\rotatebox[origin=t]{0}{good assembly}\hspace{0.55in}
    {\footnotesize\rotatebox[origin=t]{0}{bad assembly}} \\  
     \raisebox{0.50in}{\footnotesize\rotatebox[origin=t]{90}{synthetic (CAD2Render)}} \includegraphics[width=0.40\linewidth]{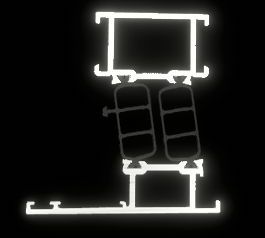}
    \includegraphics[width=0.40\linewidth]{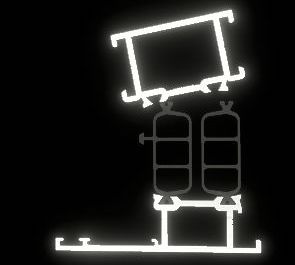} \\
    \raisebox{0.50in}{\footnotesize\rotatebox[origin=t]{90}{real examples}} 
    \includegraphics[width=0.40\linewidth]{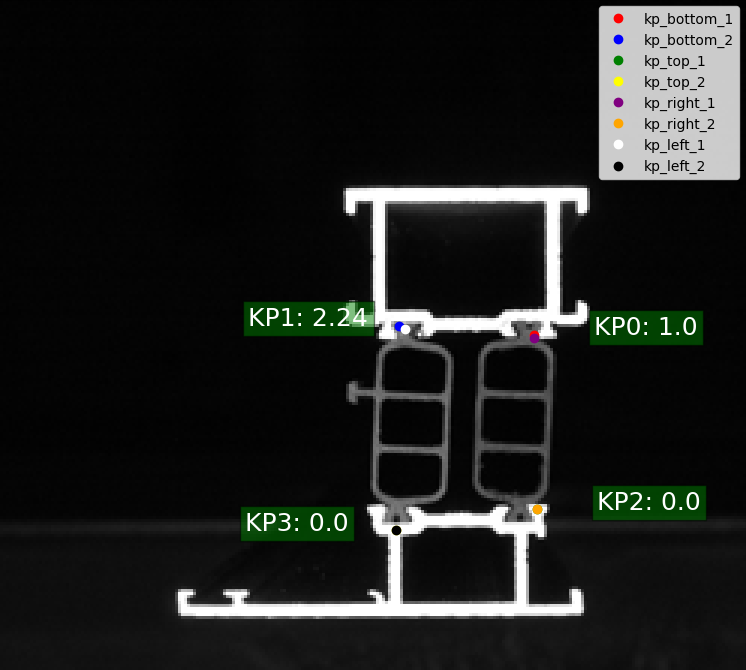}
    \includegraphics[width=0.40\linewidth]{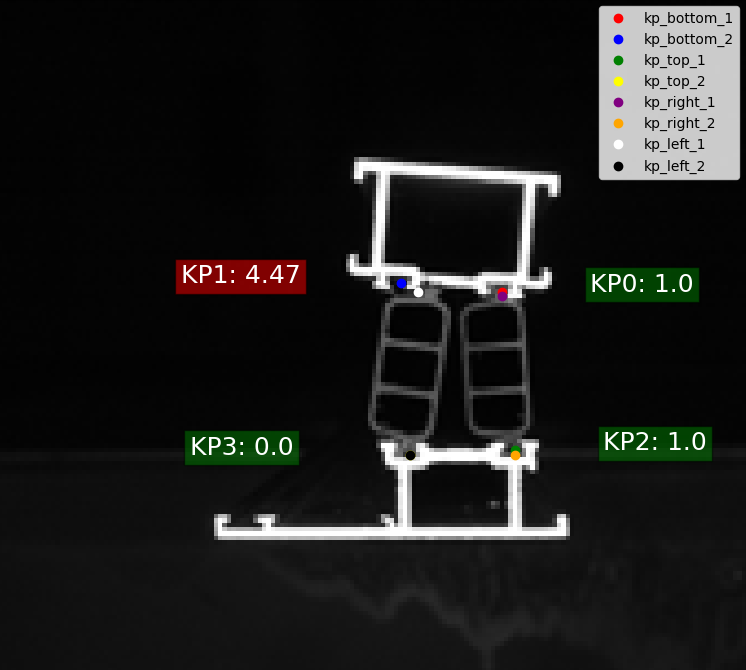}
    \caption{Assembly validation of good and bad assemblies. Top: synthetic examples generated with CAD2Render. Bottom: Output of the model shown on top of real examples in the case of a good validation (all points within spec) and a bad validation (one of the insertion areas show points that are too far from each other).}
    \label{fig:init_tool_ass}
\end{figure}

To further assess the usefulness of the keypoint detector model trained on synthetic images of CAD2Render, we applied it to an additional defect detection use case, in the form of a simple assembly validation tool. The target assembly comprises of two aluminium beams, connected by two rubber insulators. These four sections are assembled on an automated line and the assembly quality is validated using a camera with high dynamic range. Figure~\ref{fig:init_tool_ass} shows a good and a bad assembly of both real and synthetic images.

The keypoint detection model was trained for each part on the specific parts of the items that perform the insertions. During inference, the corresponding pairs for keypoints are assessed for their euclidean distance to fall below a predefined threshold. If the distance is too great, that specific insertion is assumed to be not successful and the assembly has failed. Figure \ref{fig:init_tool_ass} shows the output of the model. When applied to all input images (26 in total) the model could successfully identify good and bad assemblies in all cases.  


\section{Limitations and Future Work}
Although we performed extensive experiments showing the good performance of models trained on data generated by CAD2Render, a direct comparison to other methods is missing. Future work should compare the performance of models trained using our method to the performance of models trained on data generated by other state of the art methods. Additionally, parameters such as speed of rendering and ease of use should also be taken into account. Additionally, we would like to investigate the impact of the different quality rendering modes in CAD2render on the final model performance.
A number of training data variations were introduced to help improve generalization. A study on the impact of these variations on downstream performance is beyond the scope of this paper. For an in depth analysis of the impact of light and pose variations on object detection performance, for datasets generated by the CAD2Render tool, we refer to complementary research~\cite{object_detection_synthetic}.

\section{Conclusion}
This paper proposed a novel toolkit for synthetic data generation, that can generate a vast amount of complex photorealistic variations, including changes in lighting, appearance and pose. It is cost-effective and optimized for rendering speed on consumer hardware by exploiting the recent advancements in real-time raytracing and denoising, which is essential for fast deployment in low-volume and high-variance manufacturing. At the moment it is specifically designed for industrial use cases. However, it can be easily utilized in other domains as well, provided that there is CAD data available. Since it allows for the import and export of datasets in a standardized fashion, it can generate synthetic simulations of existing datasets, a so called digital twin. As such, it can be an enabling technology for future research on sim2real and how to close the domain gap between the real and synthetic world. Future improvements would be to include more realistic variations in the appearance of the objects based on extracted appearance of real physical examples. This would possibly reduce the sim2real domain gap that still exists in the generated data.\\

{\footnotesize
\textbf{Acknowledgement}
This research was realized in the framework of the PILS SBO project (Products Inspection by Little Supervision), funded by Flanders Make, the strategic research Centre for the Manufacturing Industry in Belgium; and the Special Research Fund (BOF, mandate ID BOF20OWB24) of Hasselt University.
}

{\small
\bibliographystyle{ieee_fullname}
\bibliography{egbib}
}

\end{document}